\documentclass[10pt,twocolumn,letterpaper]{article}

\usepackage{cvpr}              
\usepackage{amssymb}
\usepackage{pifont}
\usepackage[most]{tcolorbox} 
\usepackage{multirow}
\usepackage{array}

\newcommand{\myparagraph}[1]{\vspace{1.5pt}\noindent{\bf #1}}

\definecolor{cvprblue}{rgb}{0.21,0.49,0.74}
\usepackage[pagebackref,breaklinks,colorlinks,allcolors=cvprblue]{hyperref}

\def\paperID{11300} 
\def\confName{CVPR}
\def\confYear{2025}

\title{VideoGEM: Training-free Action Grounding in Videos}

\author{
Felix Vogel$^{1}$, Walid Bousselham$^{1,2}$, Anna Kukleva$^{3}$, Nina Shvetsova$^{1,2,3}$, Hilde Kuehne$^{1,2,4}$ \\
\small $^{1}$ Goethe University Frankfurt,
\small $^{2}$ Tuebingen AI Center/University of Tuebingen, 
\small $^{3}$ MPI for Informatics, SIC, 
$^{4}$ MIT-IBM Watson AI Lab
}

\begin{document}
\maketitle
\begin{abstract}

Vision-language foundation models have shown impressive capabilities across various zero-shot tasks, including training-free localization and grounding, primarily focusing on localizing objects in images. 
However, leveraging those capabilities to localize actions and events in videos is challenging, as actions have less physical outline and are usually described by higher-level concepts.
In this work, we propose VideoGEM, the first training-free spatial action grounding method based on pretrained image- and video-language backbones. 
Namely, we adapt the self-self attention formulation of GEM~\cite{bousselham2024grounding} to spatial activity grounding. We observe that high-level semantic concepts, such as actions, usually emerge in the higher layers of the image- and video-language models.
We, therefore, propose a layer weighting in the self-attention path to prioritize higher layers. Additionally, we introduce a dynamic weighting method to automatically tune layer weights to capture each layer’s relevance to a specific prompt.
Finally, we introduce a prompt decomposition, processing action, verb, and object prompts separately, resulting in a better spatial localization of actions. 
We evaluate the proposed approach on three image- and video-language backbones, CLIP, OpenCLIP, and ViCLIP, and on four video grounding datasets, V-HICO, DALY, YouCook-Interactions, and GroundingYouTube, showing that the proposed training-free approach is able to outperform current trained state-of-the-art approaches for spatial video grounding. \footnote{Code is available at \href{https://github.com/felixVogel02/VideoGEM}{https://github.com/felixVogel02/VideoGEM}}
\footnote{To be published at CVPR 2025. When citing this work, please refer to the final version published in IEEE Xplore. Cite as: Felix Vogel, Walid Bousselham, Anna Kukleva, Nina Shvetsova, Hilde Kuehne. “VideoGEM: Training-free Action Grounding in Videos”. In: Proceedings of the IEEE/CVF conference on computer vision and pattern recognition, 2025.}


\end{abstract}    
\section{Introduction}
\label{sec:intro}


\begin{figure}
\centering
    \includegraphics[width=0.90\linewidth]{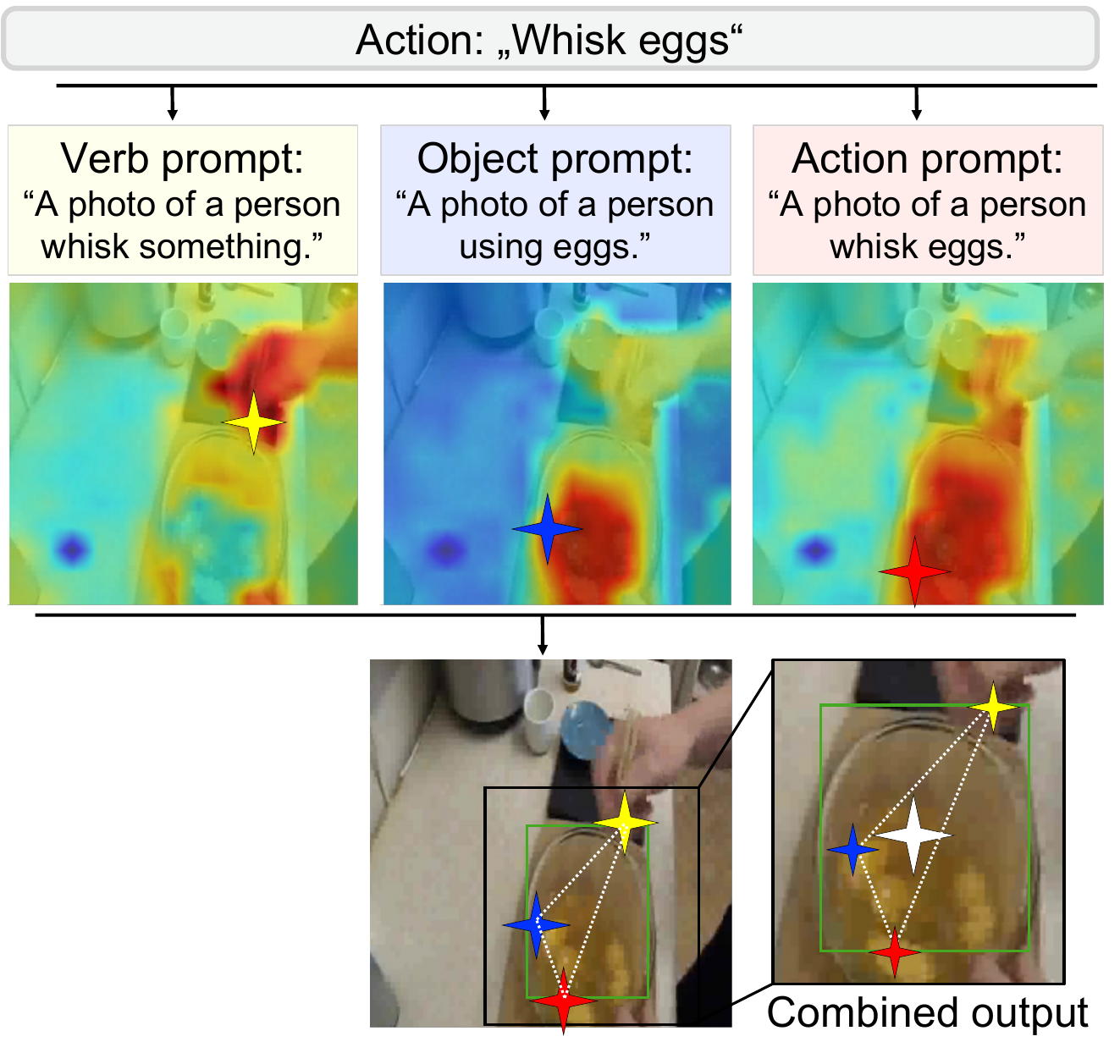    }
    \caption{\textbf{Prompt decomposition and combination.} 
    First, we decompose the action query into verb, object, and action prompts. For each component, we then predict locations corresponding to the highest values on the heatmaps (red-high, blue-low). To determine the final location (white star), we calculate the center point of these individual predictions. The red, blue, and yellow stars represent the predicted locations for the action, object, and verb prompts, respectively, while the dark green bounding box represents the annotated ground truth.
    }
    \label{fig:teaser}
    \vspace{-5mm}
\end{figure}




Spatial localization of actions in videos has been a long-standing topic in video analysis ~\cite{thi2010human, klaser2012human, gu2018ava, mettes2017spatial, soomro2018online, weinzaepfel2016human, shao2013efficient, weinzaepfel2015learning, soomro2015action}. 
While early approaches were trained with human-annotated bounding boxes and mainly focused on detecting humans and classifying respective actions~\cite{jhuang2013towards, Soomro2012UCF101}, recent approaches~\cite{chen2024and, wasim2024videogrounding, tan2021look} are trying to leverage the localization properties of vision-language models without the need of bounding box annotations, learning localization only from image- text or video-text pairs. 
However, the localization of actions based on such weak supervision presents some significant challenges: unlike objects, which usually have a clear, unambiguous outline, actions often lack such region properties and exhibit diverse, context-dependent semantics.
Additionally, actions often refer to interactions between entities and can comprise multiple objects or people over different timespans as shown in Figure~\ref{fig:teaser}. 
This variability makes it hard to capture the various visual representations of an action, especially based on web-crawled image- or video-text pairs.   
As a result, spatial video grounding methods
\cite{chen2024and, shi2019not, tan2021look} still need to be specifically trained, e.g. with a localization loss, to perform spatial localization. 


Compared to that, another line of work focuses on training-free localization of semantic concepts in images~\cite{bousselham2024grounding,li2023clip,zhou2022extract} by proposing various training-free adaptations of vision-language models to perform object localization. 
Namely, CLIPSurgery~\cite{li2023clip} proposes an alternative pathway based on value-value attention which was extended by Grounding Everything Module (GEM) \cite{bousselham2024grounding} to a self-self attention pathway. 
While those methods work well for object localization in image data, action localization requires models to capture contextual cues beyond object boundaries.
%

To address this problem, we introduce VideoGEM, the first training-free method for spatial activity grounding in videos. Inspired by GEM~\cite{bousselham2024grounding}, which focuses on spatial object localization, VideoGEM extends this approach to spatially localize activities within video content.
%
First, we extend the self-self attention pipeline to process multiple frames for action localization on video data for applying video backbones.
In this setup, self-self attention automatically spans multiple frames, aggregating attention across both space and time.
%
%
However, pretrained image-language
but also video-language backbones 
struggle to capture abstract concepts such as actions. 
To address this problem, we analyze the self-self attention pipeline of several backbones, showing that abstract concepts such as actions and verbs usually arise in higher layers. 
We, therefore, propose to weight layers of the self-self attention pipeline based on a mixture of static and dynamic weights: while static weights give more weight to higher layers, dynamic weights are adjusted based on the relevance of a layer for each prompt.

We further observe that vision-language models often show a strong object bias 
when they are prompted with verb-object combinations~\cite{yuksekgonul2023when, bousselham2024grounding, vogel2022vl}. To counteract this and focus the model on both verb and object, we propose a prompt decomposition. To this end, the \textit{verb} and \textit{object} of the \textit{action} description are extracted and separately prompted in addition to the original action. We then compute the center points of the resulting individual predictions for \textit{verb}, \textit{object}, and \textit{action}, and consider the weighted mean of all three center point predictions (see \cref{fig:teaser}). This accounts for the fact that verbs might focus more on hands, while object heatmaps automatically capture the object. 

We evaluate our approach on three pretrained backbones, CLIP \cite{radford2021learning}, OpenCLIP \cite{schuhmann2022laion}, and ViCLIP \cite{wang2023internvid, xu2021videoclip} and on four action grounding datasets, V-HICO \cite{li2021weakly}, Daly \cite{weinzaepfel2016human}, YouCook-Interactions \cite{tan2021look}, and GroundingYouTube \cite{chen2024and}.
It shows that VideoGEM allows image and video backbones to spatially ground actions in a training-free zero-shot manner and that they are able to outperform models specially trained for this setup. 
We further analyze the impact of the proposed components, including the effect of video vs image processing for self-self attention as well as the impact of layer weighting and prompt decomposition, showing how those factors contribute to the final performance.
%

We summarize our contributions as follows: 
(1) We propose VideoGEM, the first training-free method for spatial action grounding in videos that adapts self-self attention to the video domain.
(2) To capture higher-level sematic concepts such as actions, we propose a mixture of static and dynamic weights that prioritizes layers according to their relevance for capturing such concepts.
(3) We propose prompt decomposition to address the object bias in vision-language models, allowing self-self attention to consider actions, verbs, and objects independently.
(4) We show that VideoGEM outperforms even fine-tuned localization methods and provide extensive analysis of all the components. 







\section{Related Work}
\label{sec:related_work}


\myparagraph{Spatial Video Grounding.}
Spatially localizing actions in videos without the need to explicitly label respective instances has drawn significant attention in the last years. 
As one of the first works, CoMMA ( Contrastive Multilayer
Multimodal Attention)~\cite{tan2021look} proposes a multilayer cross-modal attention network that obtains an attention heatmap via attention rollout, where the predicted location is the maximum attention. 
Moreover, the authors propose the YouCook-Instructions dataset, based on YouCook2~\cite{zhou2018towards}, to evaluate spatial grounding for models pretrained on YouTube cooking data.
Compared to that, TubeDETR \cite{yang2022tubedetr} uses a space-time decoder, decoding the video-text features into a spatiotemporal object tube which includes box annotations per frame. 
STCAT \cite{jin2022embracing} proposes a Spatio-Temporal Consistency-Aware Transformer that also uses a vision and text encoder,
 followed by cross-model interaction. 
To create the object tube, the authors generate multi-modal templates to guide the decoder, followed by a prediction head. 
Recently, VideoGrounding-DINO~\cite{wasim2024videogrounding} extends GroundingDINO \cite{liu2023grounding} to videos.
Finally, What-When-Where~\cite{chen2024and} proposes a global and local loss together with a frame selection mechanism to detect actions in untrimmed videos in space and time. The authors also evaluate on a new benchmark, GroundingYouTube, which is based on MiningYouTube~\cite{kuehne2019mining}, to evaluate spatiotemporal grounding in untrimmed videos, and also use the annotated segments to evaluate spatial grounding alone.
In contrast, our method 
is training-free, using the original backbone weights without additional fine-tuning, whereas the previous methods rely on additional training of either a new projection head or the fine-tuning of the full model. 

\myparagraph{Training-free Vision-Language Grounding.}
The success of large-scale vision-language models like CLIP has generated significant interest in applying them to tasks such as open-vocabulary object localization. 
In this context, a special line of methods focuses on training-free localization \cite{zhou2022extract, li2023clip, bousselham2024grounding}, thus adapting pretrained vision-language models to handle localization tasks without changing the weights of the pretrained model.
MaskCLIP \cite{zhou2022extract} achieves this by removing the Multi-Layer Perceptron (MLP) in the vision transformer’s final layer, using the last value projection to capture dense patch-level features.
Building on this, CLIPSurgery \cite{li2023clip} introduces a parallel “surgery pathway” alongside the standard CLIP vision transformer (ViT) backbone, which operates with value-value attention instead of the usual query-key attention and connects outputs from multiple layers through residual connections.
Consistent with MaskCLIP \cite{zhou2022extract}, CLIPSurgery omits the final MLP, directly applying value-value attention. 
%
GEM \cite{bousselham2024grounding} further improves the concept of value-value attention by generalizing it to self-self attention, not only using value-value attention but also query-query and key-key attention together with a set of regularizations.
We extend the GEM model and the self-self attention mechanism by adapting it to video input and introducing a weighting 
 and a prompt decomposition mechanism, applying it to action localization.
\section{Method}
\label{sec:method}
In the following, we first review the GEM self-self attention pipeline in Section~\ref{sec:method_gem}. We then discuss its extension to videos in Section~\ref{sec:method_videogem}, the proposed layer weighting in Section~\ref{sec:method_LayerWeighting}, and the prompt decomposition in Section~\ref{sec:method_PromptDecomposition}. 
\subsection{Background (GEM)}
\label{sec:method_gem}
In the original ViT paper \cite{dosovitskiy2020image}, the attention operation is computed as follows:
\begin{equation}
\begin{aligned}
    \mathbf{A} &= \text{softmax}\left(\frac{\mathbf{X}\mathbf{W}_{q}(\mathbf{X}\mathbf{W}_{k})^T}{\tau}\right), \\
    \mathbf{O} &= \mathbf{A} \cdot (\mathbf{X}\cdot\mathbf{W}_{v}),
\end{aligned}
\end{equation}
where $\mathbf{X} = (x_i)_{i\leq N} \in \mathbb{R}^{N \times d}$ represents the patch tokens output from the ViT,
$N$ is the number of visual tokens, $d$ is the dimension of each token and  $\mathbf{W}_q, \mathbf{W}_k, \mathbf{W}_v \in R^{d \times (h * d_h)}$ are the projection matrices, where $h$ is the number of heads in the attention and $d_h$ is the dimension of each head.
%
%
%
The Grounding Everything Module (GEM) \cite{bousselham2024grounding} introduces a parallel pathway that operates alongside the original trained ViT while sharing its weights. This pathway replaces the standard self-attention mechanism with a self-self attention operation defined as follows:
\begin{equation}
\label{eq:attn_ss}
\mathbf{A}_{ss} = \text{softmax}\left( \frac{ (\mathbf{X} \mathbf{W}_{\text{proj}}) (\mathbf{X} \mathbf{W}_{\text{proj}})^\top }{ \tau } \right),
\end{equation}
where $\mathbf{W}_{proj} \in \{\mathbf{W}_q, \mathbf{W}_k, \mathbf{W}_v\}$, and $\tau$ is the temperature. 
%
The self-self attention is applied iteratively $J$ times on $L^2$ normalized visual tokens. 
For input visual tokens $\mathbf{X} \in \mathbb{R}^{N \times d}$, we denote $\mathbf{P}^{(j)}$ as the output of the self-self attention at iteration $j$: \\
%
%
\begin{equation}
    \left\{ \begin{aligned}
    \mathbf{P}^{(0)} &= \frac{\mathbf{X} \mathbf{W}_{proj}}{\Vert \mathbf{X} \mathbf{W}_{proj} \Vert_2}, \\
    \tilde{\mathbf{P}}^{(j)} &= \text{softmax}\left(\frac{\mathbf{P}^{(j-1)} (\mathbf{P}^{(j-1)})^T}{\tau}\right) \mathbf{P}^{(j-1)}, \\
    \mathbf{P}^{(j)} &= \frac{\tilde{\mathbf{P}}^{(j)}}{\Vert \tilde{\mathbf{P}}^{(j)} \Vert_2}.
\end{aligned}
\right.
\end{equation}
The final output $\mathbf{O}_{ss}$ is obtained by applying the attention matrix to the values:
\begin{equation}
    \mathbf{O}_{ss} = \text{softmax}\left( \frac{\mathbf{P}^{(J)} \cdot (\mathbf{P}^{(J)})^T}{\tau} \right) \cdot V.
\end{equation}
The final self-self attention output is obtained by averaging the query-query, key-key, and value-value attention:
\begin{equation}
    \mathbf{O}_{qkv} = \frac{(\mathbf{O}_{qq} + \mathbf{O}_{kk} + \mathbf{O}_{vv})}{3}.
\end{equation}
%
%
%
This formulation enables the patch tokens alignment with the Vision-Language Model (VLM) text encoder. Localization heatmaps can then be constructed by computing the cosine similarity between the text embedding of a prompt and the corresponding patch token representations.
%

    \begin{figure*}
    \centering
        \includegraphics[width=1\linewidth]{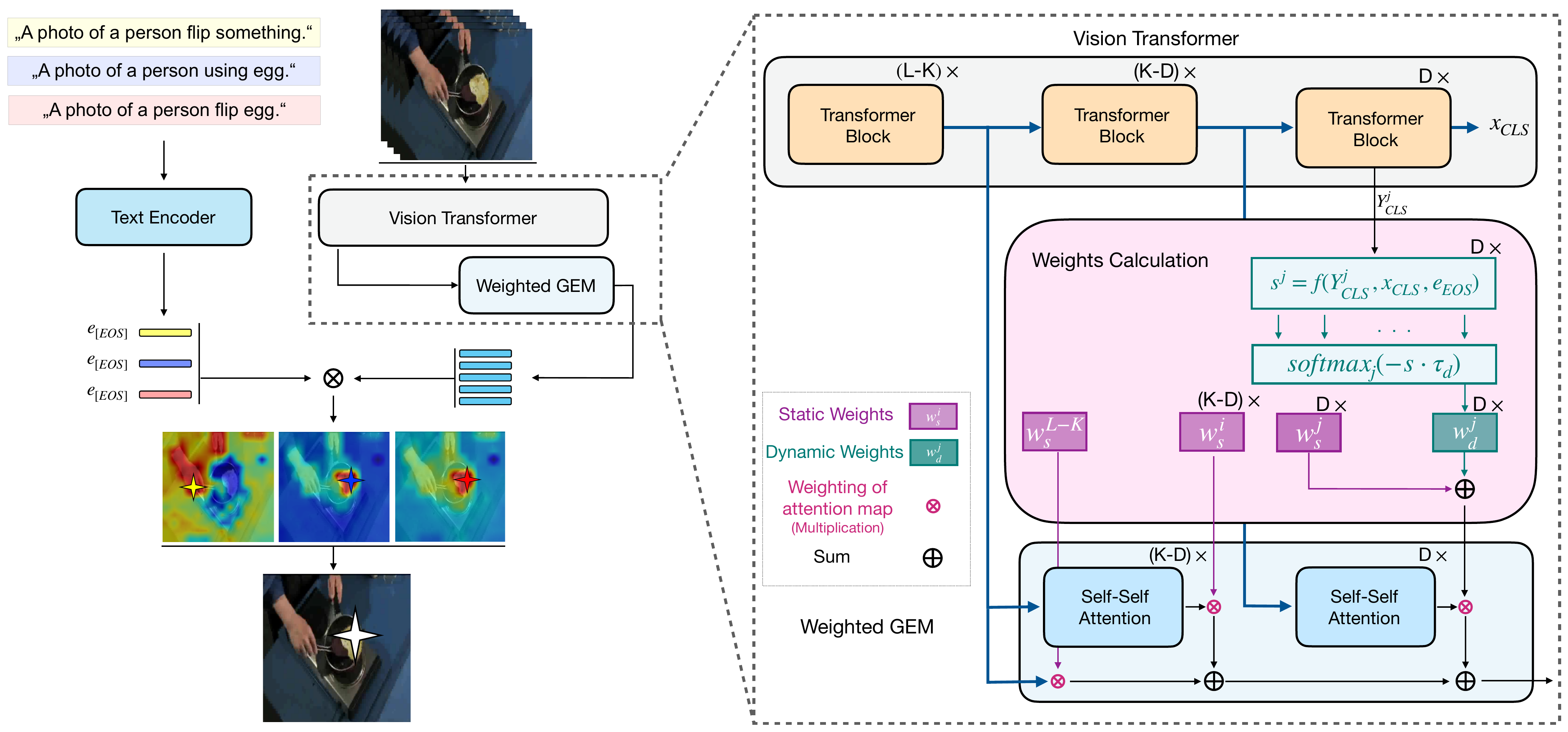}
        \caption{\textbf{Left: VideoGEM pipeline.} VideoGEM takes a video and its corresponding narration as input. Our \textit{Weighted GEM} processes the input video alongside the vision transformer to generate the representative patch tokens. Decomposition of the input narration into verb prompt, object prompt, and action prompt (see \cref{sec:method_PromptDecomposition} for details) are passed through the text encoder to obtain three \textit{[EOS]} tokens, respectively. 
        Then, three heatmaps are calculated as a similarity between patch tokens and the respective \textit{[EOS]} tokens. We then aggregate the heatmaps into one final prediction by centering the individual predicted locations. 
        \textbf{Right: Layer weighting.} 
        In our \textit{Weighted GEM} architecture, we apply a combination of static and dynamic weights. Dynamic weights are applied to the last $D$ layers, while static weights are applied to the last $K$ layers, with $K > D$. Additionally, the attention map $X^{L-K}$ is weighted by a corresponding static weight $w_s^{L-K}$. All weighted outputs from the self-attention blocks are then summed with the weighted $X^{L-K}$ attention map to produce representative patch tokens.
        The output  patch tokens of weighted GEM are used for similarity calculation with the text, resulting in an attention heatmap.
        }
        \label{fig:main}
        \vspace{-3mm}
    \end{figure*}

\subsection{GEM for Videos}
\label{sec:method_videogem}
We extend GEM to handle video inputs by adapting it to video-language models, using the ViCLIP backbone as an example. Given a video input, ViCLIP processes a sequence of $T=8$ frames $\mathbf{F} = \{f_1, ..., f_T\}$. Each frame $f_i$ is divided into $N$ patches, resulting in a total of $T \times N$ patch tokens. These tokens are processed jointly across all frames through the transformer layers, allowing the model to capture both spatial and temporal relationships.

For a given input sequence, the patch tokens from all frames are concatenated into a single sequence $\mathbf{X} \in \mathbb{R}^{(T * N) \times d}$, where $d$ is the embedding dimension. 
The self-self attention mechanism is then applied to this combined sequence of tokens. After that, we compute the cosine similarity between each patch token and the prompt embedding of the text encoder, resulting in a similarity score matrix $\mathbf{S} \in \mathbb{R}^{(T  * N)}$.
To generate the final localization output, we focus on a target frame $f_t$ positioned such that we maintain temporal context from both past and future frames. The similarity scores for frame $f_t$ are reshaped and interpolated to match the spatial dimensions of the input frame and min-max normalized to produce the final attention heatmap.
Note, that the temporal information of the video is solely considered by the video backbone.



\subsection{Layer Weighting}
\label{sec:method_LayerWeighting}
In the original GEM formulation, the self-self attention outputs are combined through residual connections, effectively giving equal importance to all layers in the final prediction. 
However, we observe that higher-level concepts, particularly actions, predominantly emerge in the higher layers of the network. 
Based on this, we propose a combination of static and dynamic weights to prioritize the contributions of different layers.
We extend the definition of $\mathbf{X}$ from the previous section, 
to $\mathbf{X}^l \in \mathbb{R}^{(T*N) \times d}$ which represents the $l$-th transformer block output, where $l \in \{1,...,L\}$ and $L$ is the total number of transformer blocks.
Let $\mathbf{Y}^l \in \mathbb{R}^{(T * N) \times d}$ denote the output of the $l$-th transformer block before the residual connection. Similarly, let $\mathbf{Z}^l\in \mathbb{R}^{(T * N) \times d}$ represent the output of the parallel self-self attention pathway at layer $l$ before the residual connection. 
Then, when applied to the last $K$ layers of the ViT, the output of GEM can be reformulated as follows:
\vspace{-1.5mm}
\begin{equation}
    \mathbf{O}_{GEM} = \mathbf{X}^{L-K}  + \sum_{l=L-K+1}^{L} \mathbf{Z}^l. \\
    \vspace{-1.5mm}
\end{equation}
\myparagraph{Static Weighting.} We first introduce static weights to assign specific importance to individual layers. For layer $l$, the static weight is defined as $w^{l}_s$ for $l \in \{L-K, ..., L\}$ resulting in the new output:
\vspace{-2mm}
\begin{equation}
    \mathbf{O}_{stat} = w^{L-K}_s \cdot \mathbf{X}^{L-K}  + \sum_{l=L-K+1}^{L} w^{l}_s \cdot \mathbf{Z}^l .
    \label{eq:static_o}
    \vspace{-2mm}
\end{equation}
Practically, $w^{l}_s$ increases monotonically with $l$.

\myparagraph{Dynamic Weighting.} We further introduce dynamic weights that adapt to the semantic requirements of each input prompt by analyzing each layer's contribution to the model's understanding of the prompted concept.

Let $\mathbf{x}_{CLS} = \mathbf{X}_{0}^L \in \mathbb{R}^d$ denote the [CLS] token representation from the final layer, which can be decomposed into the sum of residuals from all layers: $\mathbf{x}_{CLS} = \sum_{l=1}^L \mathbf{Y}_{0}^l = \sum_{l=1}^L \mathbf{Y}_{CLS}^l$. The text embedding of the prompt obtained from the text encoder is defined as $\mathbf{e}_{EOS} \in \mathbb{R}^d$. We measure each layer's importance by evaluating how its removal affects the alignment between visual and textual representations, for details see~\cref{sec:eval_ablation}. Let $s = (s^{L-D+1}, ..., s^L)$ be the similarity vector where we compute the similarity score for layer $l$ as:
%
\vspace{-1.5mm}
\begin{equation}
    s^l = \frac{(\mathbf{x}_{CLS} - \mathbf{Y}_{CLS}^l) \cdot \mathbf{e}_{EOS}}{\Vert \mathbf{x}_{CLS} - \mathbf{Y}_{CLS}^l \Vert_2 \cdot \Vert \mathbf{e}_{EOS} \Vert_2},
    \vspace{-1.5mm}
\end{equation}
The dynamic weights for the last $D$ layers with $D \leq K$ are computed using the similarity vector $s$ through a softmax operation with temperature $\tau_d$. 
We define the dynamic weight for layer $l \in\{L-D+1, ..., L\}$ as:
%
\vspace{-1mm}
\begin{equation}
    w^{l}_d = \text{softmax}_{l}(-s \cdot \tau_d),
\label{eq:dynamic}
\vspace{-1mm}
\end{equation}
%
then the final output for dynamic weights is computed as:
%
\vspace{-1.5mm}
\begin{equation}
    \mathbf{O}_{dyn} =  \mathbf{X}^{L-K}+\sum_{l=L-K+1}^{L-D}\mathbf{Z}^l+\sum_{l=L-D+1}^{L} w^{l}_d\cdot\mathbf{Z}^l .
    \label{eq:dynamic_o}
\end{equation}
It is important to note that the [CLS] token representations used for computing similarities are extracted from the original self-attention pathway rather than the self-self attention. It is motivated by the fact that the [CLS] token in the original model was specifically trained to align with the text encoder's [EOS] token representations during pre-training.


\myparagraph{Combining Static and Dynamic Weights.}
To leverage the complementary benefits of both weighting schemes, we propose a unified approach that combines static and dynamic weights (see \cref{fig:main}). For a network with $L$ layers, we apply static weights to the last $K$ self-self attention layers and the previous self attention input while computing dynamic weights for the last D layers (where $D \leq K$). The combined weights $w^{l}_c$ for $l \in \{L-K, ..., L\}$ are defined as:
\vspace{-2mm}
\begin{equation}
w^{l}_c =
    \left\{ \begin{aligned}
    & w^{l}_s - \frac{1}{D} + w^{l}_d & \text{if } l > L-D \\
    &w^{l}_s & \text{otherwise}
    \label{eq:weights_comb}
    \vspace{-2mm}
\end{aligned}
\right.
\end{equation}
%
where $w^{l}_s$ denotes static weight and $w^{l}_d$ represents the dynamic weight as defined in~\cref{eq:dynamic}.
We subtract $\frac{1}{D}$ from static weights when dynamic weights are applied to not increase the sum of weights.
The final output incorporating these combined weights is computed as:
%
\vspace{-2mm}
\begin{equation}
    \mathbf{O}_{comb} = w^{L-K}_c \cdot \mathbf{X}^{L-K} + \sum_{l=L-K+1}^{L} w^{l}_c \cdot \mathbf{Z}^l 
    \label{eq:weights_comb_applied}
    \vspace{-2mm}
\end{equation}
This weighted combination allows the model to adaptively balance the static prior knowledge about layer importance with dynamic, prompt-specific adjustments, particularly in higher layers where semantic concepts emerge.
\subsection{Prompt Decomposition for Action Grounding}
\label{sec:method_PromptDecomposition}

Action descriptions typically consist of two key components: A verb describing the action itself and one or more objects involved in the action. To effectively leverage this inherent structure, we propose a prompt decomposition method that processes these components separately while maintaining the context of the complete action description.
Namely, we decompose each action query into three distinct components: a verb prompt, an object prompt, and the original action prompt. For consistent processing, we employ the following template format:
\begin{tcolorbox}[title=Prompt Templates]
\begin{itemize}
    \item Verb: \textit{A photo of a person \{verb\} something.}
    \item Object: \textit{A photo of a person using \{object\}.}
    \item Action: \textit{A photo of a person \{action prompt\}.}
\end{itemize}
\end{tcolorbox}
In case a component, verb or object, is missing or cannot be extracted, we use fallback templates, \textit{"A photo of a person doing something."} for missing verbs and \textit{"A photo of a person."} for missing objects. 
%
For each prompt, we compute the similarity between its text embedding and the visual tokens,
identifying the regions of highest activation. The model then determines a center point for each attention map, corresponding to the location with maximum similarity between the text and visual representations.

To combine these separate predictions into a final localization output, we employ a weighted averaging scheme that prioritizes the action prompt while incorporating information from the component-specific predictions (see \cref{fig:teaser,fig:main}). Let $c_{verb}, c_{obj}, c_{act} \in \mathcal{R}^2$ denote the predicted center coordinates for the verb, object, and action prompts respectively, and $w_{verb}, w_{obj}, w_{act} \in \mathcal{R}$ 
their corresponding weights, where we assign higher weights to action prompts. The final prediction $c_{dec}$ is computed as:
\vspace{-2mm}
\begin{equation}
    c_{dec} = w_{verb} \cdot c_{verb} + w_{obj} \cdot c_{obj} + w_{act} \cdot c_{act}
    \label{eq:prompt_decomp}
    \vspace{-2mm}
\end{equation}
The weighted combination serves two purposes. First, it maintains the primacy of the action prompt's prediction while allowing refinement based on component-specific localizations. Second, if any single component prediction deviates from the true action center, the ensemble nature of the prediction helps maintain accuracy through the influence of the other components. This is particularly valuable for complex actions where the verb and object locations might provide complementary spatial information.

\myparagraph{Extracting Verbs and Objects.}
\label{sec:extract_verbs_objs}
For datasets with a label structure of \textit{"verb\_object"} verb and object can be directly retrieved.
For natural language annotations as in YouCook-Interactions, we select the verbs and objects as the ones that are most likely visible in the input. 
To this end, we extract all verbs and objects from the action description with a natural language processing (NLP) tool. 
We then generate an individual prompt for each verb and object and apply the respective vision-language backbone to determine which verb and object show the highest similarity to the input.

\begin{table*}
\centering
    \resizebox{1\linewidth}{!}{
    \begin{tabular}{ 
    p{4.05cm}p{2.6cm}p{3.8cm}p{3.8cm}|p{0.65cm}p{0.65cm}p{0.65cm}p{0.65cm}|p{0.65cm} 
    }
     \toprule
     Model & Backbone & Backbone Pretraining Data & Grounding Training Data & VH & Daly & YC & gYT & Avg. \\
     \midrule
     \multicolumn{3}{l}{\textbf{Models trained for grounding with localization supervision:}} \\
     TubeDETR\cite{yang2022tubedetr}\textdagger & ResNet101\cite{he2016deep},  RoBERTa\cite{liu2019roberta} & ImageNet, Visual Genome, Flickr, COCO &  
     VidSTG & - & - & 51.63 & -& -\\      
     STCAT~\cite{jin2022embracing}\textdagger & ResNet101\cite{he2016deep},  RoBERTa\cite{liu2019roberta} & - & VidSTG &- & - & 55.90 & -& -\\  
     VideoGrounding-DINO\cite{wasim2024videogrounding} & Swin-L; BERT\cite{devlin2018bert} & O365, OI, GoldG, Cap4M, COCO, RefC
     & VidSTG &- & - & 57.73 & -& -\\
     GLIP\cite{li2022grounded}\textdaggerdbl & Swin-L* \cite{liu2021swin} & - & FourODs,GoldG,Cap24M&66.05 & - & 52.84 & 53.62 & -\\
     \midrule
     \midrule
    \multicolumn{3}{l}{\textbf{Models trained for grounding with vision-text pairs only:}} \\
     CoMMA\cite{tan2021look}\textdaggerdbl & CLIP &  WIT-400M  & HT100M & 55.20 & 61.06 & 52.65 & 47.56 & 54.12\\
     RegionCLIP\cite{zhong2022regionclip}\textdaggerdbl & RN50x4* &  WIT-400M  & CC3M & 57.92 & 67.12 & 51.56 & 52.84 & 57.36\\
     WWW-CLIP\cite{chen2024and} & CLIP &  WIT-400M  & HT100M & 60.71 & 70.08& 57.10 & 55.49 & 60.85 \\
     WWW-CLIP\cite{chen2024and} & CLIP* &  WIT-400M  & HT100M & 62.34 & 71.35& 58.35& 56.98 & 62.26\\
     \midrule
     \midrule
     \multicolumn{3}{l}{\textbf{No training for grounding:}} \\
     \multirow{3}{*}{GEM\cite{bousselham2024grounding}}
     & CLIP &  WIT-400M & - & 67.79 & 78.52 &50.08 & 36.92 & 58.33\\
      & OpenCLIP & LAION2B & - & 68.28 & 74.05 &56.87 & 32.91 & 58.03\\
     & ViCLIP & InternVid-FLT-10M & - & 65.08 & 73.75 & 53.62 & 51.28 & 60.93\\
     \midrule
     \multirow{3}{*}{VideoGEM (ours)}
     &  CLIP & WIT-400M & -  &\textbf{76.90} & \textbf{84.53} & 52.57 & 47.46 & 65.37 \\
       & OpenCLIP & LAION2B & - & 76.42 & 80.32 &\textbf{60.05} & 45.33 & 65.53\\
     & ViCLIP & InternVid-FLT-10M & - & 75.75 & 78.25 & 55.10& \textbf{57.21} & \textbf{66.58}\\
     \bottomrule
    \end{tabular}  
    }
\caption{\textbf{Accuracy of VideoGEM compared to the State-of-the-Art.} VideoGEM includes prompt decomposition, and static and dynamic weights. GEM is applied with the same action prompt as VideoGEM. We compare the accuracy on V-HICO (VH), Daly, YouCook-Interactions (YC), and GroundingYouTube (gYT). Finetuned backbones are marked with *. \textdagger  results from  \cite{wasim2024videogrounding}. \textdaggerdbl  results from \cite{chen2024and}.   }
\label{table:final_results}
\vspace{-4mm}
\end{table*}

\section{Results}
\label{sec:evaluation}
\subsection{Datasets}
\label{sec:eval_datasets}
We evaluate four video datasets for action grounding. Since our proposed method is training-free, we only consider the test set of the datasets for our evaluation, if there is one. 

\myparagraph{V-HICO} (Videos of Humans Interacting with Common Objects) \cite{li2021weakly} has a test set containing 608 videos with annotated bounding boxes for a human performing an action and the object on which the action is performed. It consists of 244 object classes and 99 action classes with a total of 756 action-object pairwise classes. Following the evaluation of \cite{chen2024and} we use the union of the human and object bounding boxes as ground truth for V-HICO. 

\myparagraph{DALY} (Daily Action Localization in YouTube videos) \cite{weinzaepfel2016human} annotates 510 YouTube videos with ten action classes. 

\myparagraph{GroundingYouTube} \cite{chen2024and} is based on the MiningYouTube dataset \cite{kuehne2019mining} which contains 250 cooking videos from YouTube. GroundingYouTube provides spatio-temporal annotations including bounding boxes for actions.

\myparagraph{YouCook-Interactions} \cite{tan2021look} is based on the validation set of YouCook2 \cite{zhou2018towards} that contains 457 videos. It provides spatial bounding box annotations for interactions with labels as natural language sentences.



\subsection{Implementation Details}
\label{sec:eval_setup}

We use GEM with $K=7$ self-self attention layers and $J=1$ iterations according to GEM \cite{bousselham2024grounding}, using the static weights: $w_{s} = [0.3, 0.4, 0.5, 0.6, 0.7, 0.9, 0.9, 0.9]$ for the seven self-self attention layers, and its initial self attention input. Note that $w_{s}$ starts at index $L-K$.
We further apply dynamic weights for the last $D=3$ layers. We use a temperature of $\tau_d=20$ for dynamic weights according to Equation \ref{eq:dynamic} for all datasets and models.
For prompt decomposition we apply the weights $w_{verb} = 0.2, w_{obj} = 0.2, w_{act} = 0.6$ according to Equation \ref{eq:prompt_decomp}.
We evaluate on ViCLIP, OpenCLIP, and CLIP as backbones in a training-free manner meaning that they are not specifically trained for action localization. %
For ViCLIP, we sample $7$ frames around the labeled frame to get a video input. $4$ frames are sampled before, $3$ frames are sampled after the labeled frame.
Accuracy is used as the main evaluation metric. A prediction is correct if the predicted location is inside the ground-truth bounding box, otherwise it is false. The accuracy is calculated as the proportion of the correct predictions.
%
%
%
\begin{figure}[t]
\centering
    \includegraphics[width=0.99\linewidth]{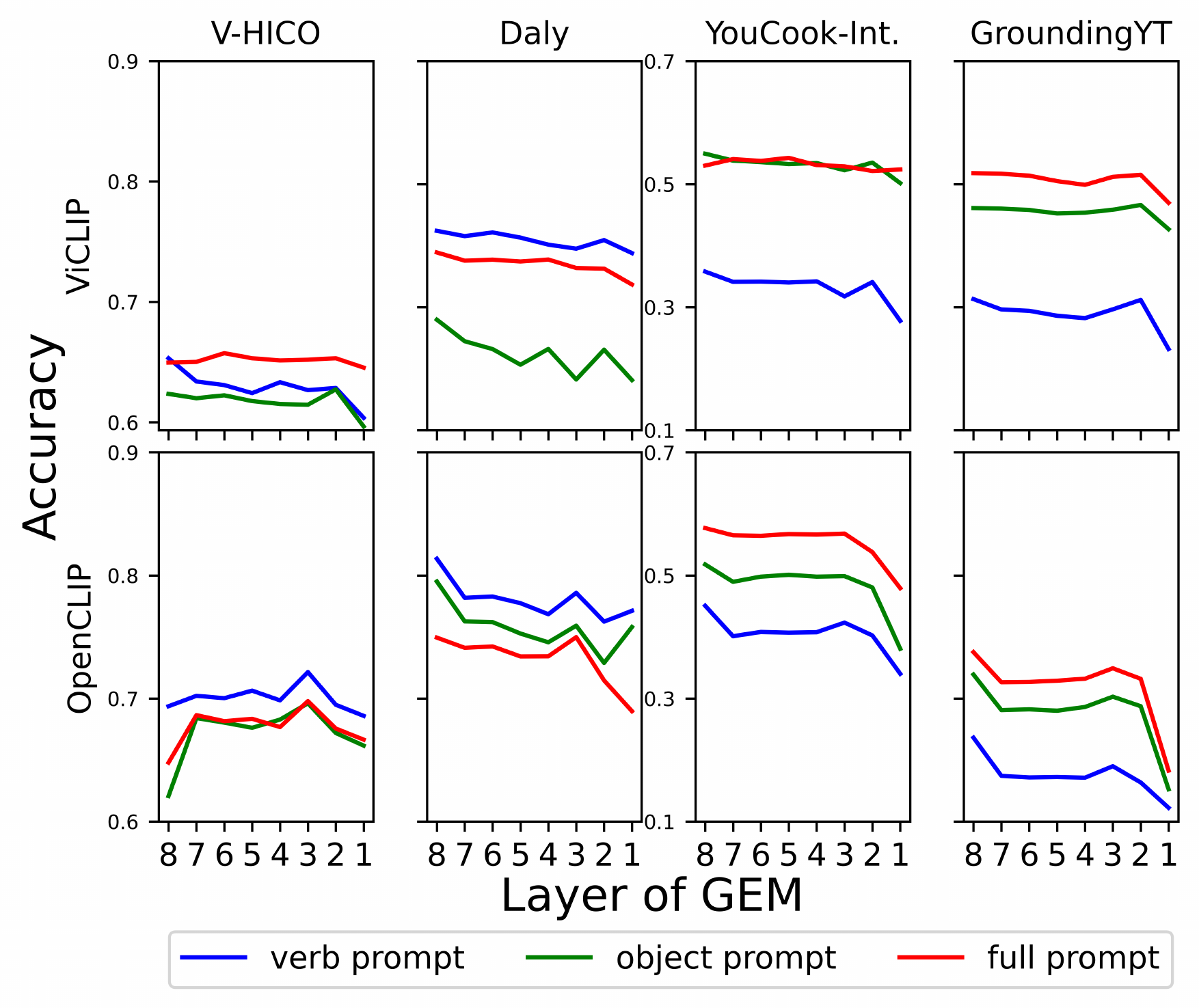}
    \caption{\textbf{Importance of GEM layers.} 
    The accuracy of GEM with one removed layer is calculated. The removed layer index is on the x-axis where $1$ is the final layer of GEM going down to $8$ which is the initial self attention input to GEM. }
    \label{fig:gem_importance}
    \vspace{-5mm}
\end{figure}
\begin{figure}[t]
\centering
    \includegraphics[width=0.99\linewidth]{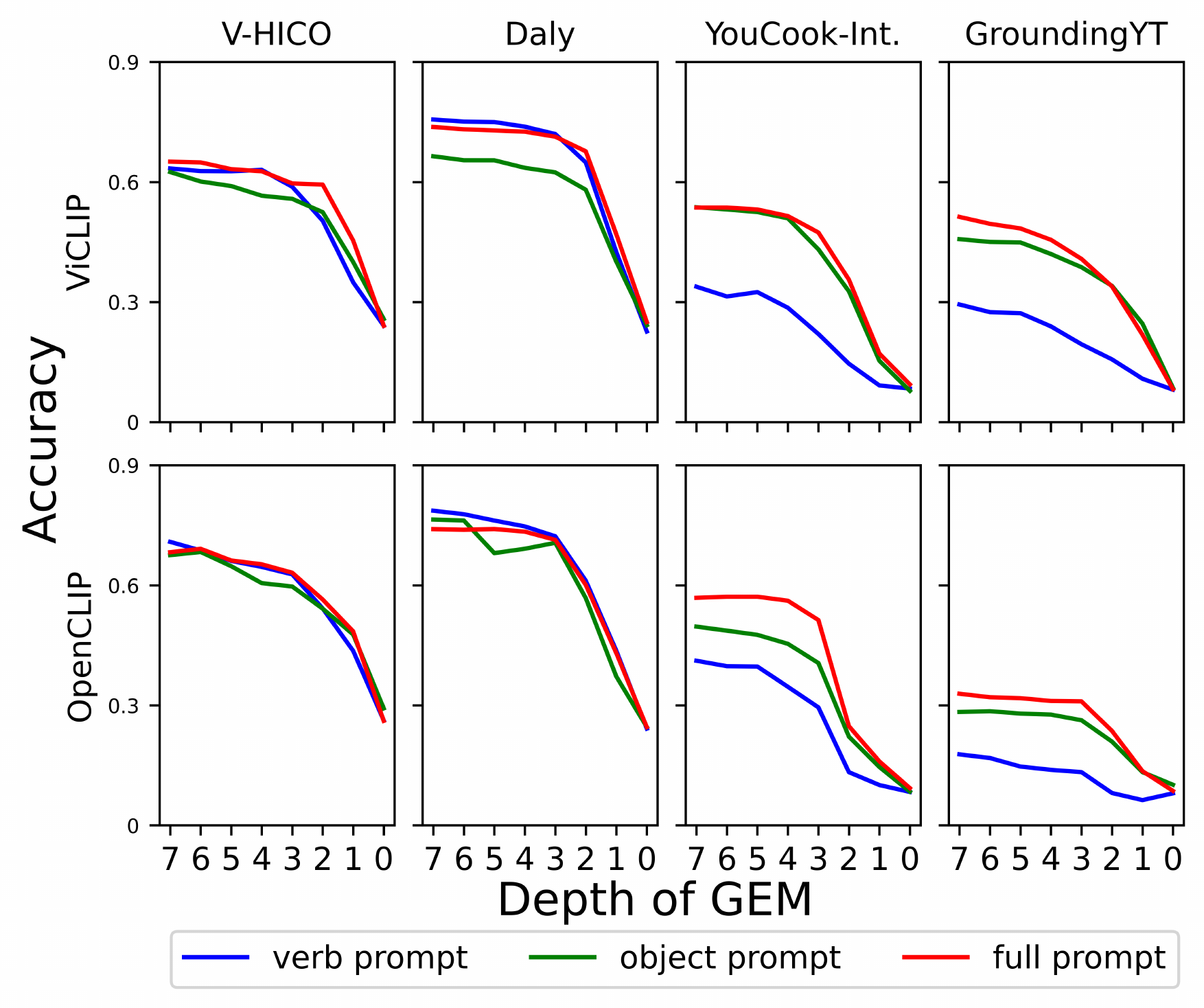}
    \caption{\textbf{Influence of the number of GEM layers.}
    Up to seven layers are added for GEM starting with a self-self attention layer for the final Transformer block.
    With zero layers, the output equals to the output of the backbone without GEM.}
    \label{fig:gem_depth}
    \vspace{-5mm}
\end{figure}
%
%
%
\subsection{Comparison to State-of-the-Art}
\label{sec:eval_sota}
We first compare the proposed approach to trainable video grounding methods as well as to the vanilla GEM approach in \cref{table:final_results}. To the best of our knowledge, all existing video grounding models are either specifically trained for action grounding or have a fine-tuned backbone for action localization, whereas our setup is training-free.

First, we show that on average our VideoGEM outperforms all other methods by more than 3$\%$ with any of the three backbones.
%
Looking at the different backbones in detail, we notice that especially CLIP and OpenCLIP, so backbones specialized for objects, perform well on the V-HICO and Daly datasets, while ViCLIP significantly outperforms all other backbones on GroundingYouTube. We attribute this to the fact that V-HICO and Daly focus mainly on object annotations, while GroundingYouTube mainly focuses on annotations based on the center point of the action, therefore, stronger deviates from object-centered bounding boxes. Moreover, YouCook-Interactions and GroundingYouTube only contain videos from the specialized cooking domain, while V-HICO and Daly contain more general actions. General actions might be easier recognizable in a zero-shot object-centric setting without specific domain knowledge.
%
Compared to the original GEM pipeline, we show that VideoGEM always improves over GEM independently of the dataset and the backbone that is chosen.
%
%
%
%

\subsection{Ablations}
\label{sec:eval_ablation}

\myparagraph{Layer Importance.}
\cref{fig:gem_importance} we first visualize the per-layer impact by excluding single layers from one to eight layer from GEM. We observe that the accuracy is lower when the final layers of GEM are excluded compared to excluding early layers.
This might lead to the assumption, that reducing the number of GEM layers would boost performance as well.
Compared to that however, \cref{fig:gem_depth} shows the results when adding up to seven layers with zero layers corresponding to the
output of the backbone without GEM. It shows that the performance of GEM depends on the overall number of self-self attention layers and that the accuracy increases for more layers saturating at a depth of seven that we use for all our experiments, also suggested by the original GEM paper \cite{bousselham2024grounding}. This strengthens the concept of static weights, suggesting that higher layers generally capture complex concepts better than earlier layers, justifying why they should be weighted higher in the final decision. Earlier layers on the other hand can still capture important concepts such that they should not be excluded entirely, but just given less importance for the final decision.

\newcolumntype{P}[1]{>{\centering\arraybackslash}p{#1}}

\begin{table}
\centering
    \begin{tabular}{ 
    P{1.05cm}P{1cm}|P{0.65cm}P{0.65cm}P{0.65cm}P{0.65cm}|P{0.65cm}  }
    
     \hline
     Backbone & Weights  & VH & Daly & YC & gYT & avg \\
     \hline

        \multirow{4}{*}{ViCLIP}
       & none & 74.79 & 76.84 & 54.38 & 56.39 & 65.60\\
        & dyn& 74.49 & 76.85 & 54.62 & 56.47 & 65.61\\
       & stat& \textbf{76.18} & \textbf{78.38} & 55.02& 56.75 & \textbf{66.58}\\
       & s+d& 75.75 & 78.25 & \textbf{55.10}& \textbf{57.21} & \textbf{66.58}\\
       \hline

       \multirow{4}{*}{OpenCLIP}
        & none&77.86 & 79.27 & 59.20 & 40.17 & 64.13\\
       & dyn& \textbf{78.41} & 79.07 & 59.77 & 43.29 & 65.14 \\
       & stat& 76.12 & 80.30 & \textbf{61.82} & 42.96 & 65.30\\
       & s+d& 76.42 & \textbf{80.32} & 60.05 & \textbf{45.33} & \textbf{65.53} \\
       \hline
    \end{tabular} 
\caption{\textbf{Influence of different layer weighting strategies}. VideoGEM with prompt decomposition is applied without layer weighting (none), only static (stat), only dynamic (dyn), or combined weights (s+d) on V-HICO (VH), Daly, YouCook-Interactions (YC), and GroundingYouTube (gYT).}
\label{table:weights}
\vspace{-5mm}
\end{table}
\myparagraph{Layer Weighting.}
%
%
To assess the effect of different weighting strategies, we compare static, dynamic, and combined weights (Equation \ref{eq:static_o}, \ref{eq:dynamic_o}, and \ref{eq:weights_comb_applied}) with using no weights in~\cref{table:weights} observing a performance increase by about  $1\%$ on average for both backbones compared to no layer weighting. We observe that the effect of dynamic weighting depends on the backbone as it relies on representative \textit{[CLS]} tokens for the final layers. 
If the \textit{[CLS]} token is primarily formed in the last layer, the benefit of dynamic weights is reduced, as seen with the ViCLIP backbone. However, with OpenCLIP, we observe consistent improvements. Notably, on the Grounding YouTube dataset as the most distinct from other object-centric datasets, dynamic weights boost performance by more than $3\%$.
We provide further insights in the supplementary material in \cref{table:verb_weights,table:object_weights,table:action_weights}.

\newcolumntype{P}[1]{>{\centering\arraybackslash}p{#1}}

\begin{table}
\centering
    \begin{tabular}{ 
    P{1.3cm}P{0.75cm}|P{0.65cm}P{0.65cm}P{0.65cm}P{0.65cm}|P{0.65cm}  }
    
     \hline
     Backbone & Mode  & VH & Daly & YC & gYT & avg \\
     \hline
    
    \multirow{4}{*}{ViCLIP}
       & verb& 64.72  & 76.21& 36.54&31.26 & 52.18 \\
       & obj &62.30 & 68.71 & 54.94 & 46.78 & 58.18\\
       & act &  65.68 & 74.17  & 52.97 & 51.99 & 60.95\\
       & all & \textbf{75.75} & \textbf{78.25} & \textbf{55.10}& \textbf{57.21} & \textbf{66.58}\\
       \hline

       \multirow{4}{*}{OpenCLIP}
        & verb& 69.60 & 80.24 &  43.45 & 21.60 & 53.72\\
       & obj & 65.44 & 77.00 & 51.21 &  34.20 & 56.96 \\
       & act &  66.77 & 74.87 &  57.36 & 38.38 & 59.35\\
       & all & \textbf{76.42} & \textbf{80.32} & \textbf{60.05} & \textbf{45.33} & \textbf{65.53}\\
     \hline
    \end{tabular} 
\caption{\textbf{Influence of prompt decomposition.}
VideoGEM with static and dynamic weights is applied without prompt decomposition to only verb, object (obj), or action (act) prompts. The results are compared to VideoGEM with static and dynamic weights as well as prompt decomposition (dec) on V-HICO (VH), Daly, YouCook-Interactions (YC), and GroundingYouTube (gYT).}
\label{table:prompt_decomp}
\vspace{-2mm}
\end{table}
\newcolumntype{P}[1]{>{\centering\arraybackslash}p{#1}}

\begin{table}
\centering
    \begin{tabular}{ 
    P{1.3cm}P{0.75cm}|P{0.65cm}P{0.65cm}P{0.65cm}P{0.65cm}|P{0.65cm}  }
    
     \hline
     Backbone & Mode  & VH & Daly & YC & gYT & avg \\
     \hline
     \multirow{4}{*}{ViCLIP}
        & none & 65.68 & 74.17 & 52.97 & 51.99 & 60.95\\
       & mul & 66.16 & 76.72 & 55.91 & 48.87 & 61.92\\
       & avg& 65.44 & 75.99 & \textbf{56.75} & 49.96 & 62.04\\
       & ours& \textbf{75.75} & \textbf{78.25} & 55.10& \textbf{57.21} & \textbf{66.58}\\
       \hline
       
    \multirow{4}{*}{OpenCLIP}
       & none & 66.77 & 74.87 &57.36 & 38.38 & 59.35\\
        & mul& 68.28 & \textbf{80.36} & 58.40 & 34.18 & 60.31\\
       & avg& 68.40 & 78.94 & 58.04 & 35.34 & 60.18\\
       & ours&\textbf{76.42} & 80.32 & \textbf{60.05} & \textbf{45.33} & \textbf{65.53} \\
       \hline
    \end{tabular} 
\caption{\textbf{Influence of merging strategies for prompt decomposition.} VideoGEM (ours) combines verb, object, and action prompt predictions with a weighted average of the predicted positions. This combination technique is compared to the element-wise multiplication (mul) and element-wise averaging (avg)  of heatmaps before taking the highest attention value of the resulting heatmap as prediction. \textit{None} corresponds to the baseline evaluation without prompt decomposition for action prompts with combined weights. We use V-HICO (VH), Daly, YouCook-Interactions (YC), and GroundingYouTube (gYT) for testing.}
\label{table:prompt_merg}
\vspace{-5mm}
\end{table}

\myparagraph{Prompt Decomposition.} 
We evaluate the effect of prompt decomposition in \cref{table:prompt_decomp} by applying VideoGEM without prompt decomposition for only a verb, object, or action prompt, compared to our proposed prompt decomposition method including all prompts.
Averaged over all datasets, using prompt decomposition improves over using only a single (verb, object, or action-) prompt by over $5\%$ independently of the used backbone.
Each prompt alone achieves strong performance, supporting the approach of first independently processing prompts and only then aggregating their individual predictions. Our method demonstrates their compatibility for recognizing complex activities effectively.

Additionally, we compare different merging strategies in \cref{table:prompt_merg} for obtaining the final prediction given the three attention heatmaps for the verb, object, and action-prompt.
VideoGEM that averages the predicted positions, is evaluated against averaging or multiplying heatmaps elementwise. The ratio of weights for the verb, object, and action prompt is always $1:1:3$ independent of the merging strategy.
%
Compared to standard GEM, all merging strategies improve significantly.
That suggests that decomposing the prompt into its relevant parts enforces the model to focus on every important part compared to only using the full prompt where the model can neglect the verb and mainly focus on the object ~\cite{yuksekgonul2023when, bousselham2024grounding, vogel2022vl}.
Moreover, averaging positions boosts performance much further compared to merging heatmaps (additively, or multiplicatively). 
This can be explained by two effects. First, it centers the action by predicting a position between the main parts of the action making it more robust. Second, it has self-correcting abilities. If one prediction is slightly off, the other predictions can drag the wrong prediction back to the others. It can thus be seen as an ensemble model having three votes for the correct prediction instead of just one, improving robustness and accuracy.

\newcolumntype{P}[1]{>{\centering\arraybackslash}p{#1}}

\begin{table}
\centering
    \begin{tabular}{ 
    P{1.5cm}P{0.55cm}|P{0.65cm}P{0.65cm}P{0.65cm}P{0.65cm}|P{0.65cm}  }
    
     \hline
     Model & Data & VH & Daly & YC & gYT & avg \\
     \hline
    \multirow{2}{*}{GEM}
       & vid & 65.08 & 73.75 & \textbf{53.62} & \textbf{51.28} & \textbf{60.93} \\
       & img & \textbf{65.20} & \textbf{74.00} & 52.17 & 48.80 & 60.04\\
       \hline

    \multirow{2}{*}{VideoGEM}
      & vid & \textbf{75.75} & 78.25 & \textbf{55.10} & \textbf{57.21} & \textbf{66.58}\\
      & img & 74.19 & \textbf{78.47} & 54.86 & 55.08 & 65.65\\

     \hline
    \end{tabular} 
\caption{\textbf{Influence of video input.} VideoGEM with ViCLIP as backbone is evaluated on image and video inputs. For reference we also evaluate GEM with ViCLIP as a backbone. 
We evaluate on V-HICO (VH), Daly, YouCook-Interactions (YC), and GroundingYouTube (gYT).
}
\label{table:viclip_results}
\vspace{-4mm}
\end{table}
\myparagraph{Image vs. Video Data.}
To determine the importance of video data for action grounding, we compare ViCLIP for video and image input in~\cref{table:viclip_results}.
ViCLIP takes eight images as input. In the image setting, we give the same image repeated eight times as an input.
For the video setting, we use subsequent frames as input according to \cref{sec:eval_setup}.
Using video input outperforms its image-based counterpart independently of using standard GEM or VideoGEM by almost $1\%$ on average.
While the video input increases accuracy on YouCook-Interactions and GroundingYouTube compared to the image input, it performs similarly on V-HICO and Daly.
This can be attributed to the more static actions in V-HICO and Daly like \textit{"snapping fingers"} for V-HICO or \textit{"Drinking"} for Daly, while f.e. \textit{"chopping"} or \textit{"arranging"} in cooking videos as in YouCook-Interactions or GroundingYouTube is more dynamic.
Note, that the only difference between the video and image input is, that the image input repeats the same image $8$ times, while for video input surrounding frames are used. If the actions are static and the surrounding frames are very similar, the video and image input are also very similar resulting in only minor performance differences.


\section{Conclusion}
\label{sec:conclusion}
In this work, we introduced VideoGEM, the first training-free method for action grounding in videos. We proposed a weighting technique using static and dynamic weights to assign greater importance to layers that capture conceptually relevant information for complex activities. Additionally, we introduced prompt decomposition to fully leverage action prompts, helping to reduce object bias in standard image- and video-language models. Remarkably, our training-free VideoGEM outperforms all previous state-of-the-art methods that rely on fine-tuning backbones for action localization tasks.

\section{Acknowledgments}
{\small 
Walid Bousselham is supported by German Federal Ministry of Education and Research (BMBF) project STCL - 01IS22067. 
Nina Shvetsova is supported in part by German Federal Ministry of Education and Research (BMBF) project STCL - 01IS22067. 
Prof. Hilde Kuehne is supported in part by the ERC Starting Grant GraViLa 101117556. 
We thank Dr. Brian Chen for his support and helpful discussions.
}

{
    \small
    \bibliographystyle{ieeenat_fullname}
    \bibliography{main}
}
\clearpage
\setcounter{page}{1}
\maketitlesupplementary

%
The supplementary material is organized as follows: first, we show a comparison between GEM and our proposed weighting mechanism in \cref{sec:weighted_gem_vs_gem}.
Then, we extend our evaluation to include the BLIP backbone~\cite{li2022blip} in \cref{sec:blip}. Next, we present additional experiments on extending image models to video data in \cref{sec:video} and offer a more detailed analysis of the effects of static weights, dynamic weights, and their combination in \cref{sec:weights_abl_suppl}, and finally, we provide qualitative evaluation in \cref{sec:qual_suppl}.
    \begin{figure*}
\begin{center}
        \includegraphics[width=0.99\linewidth]{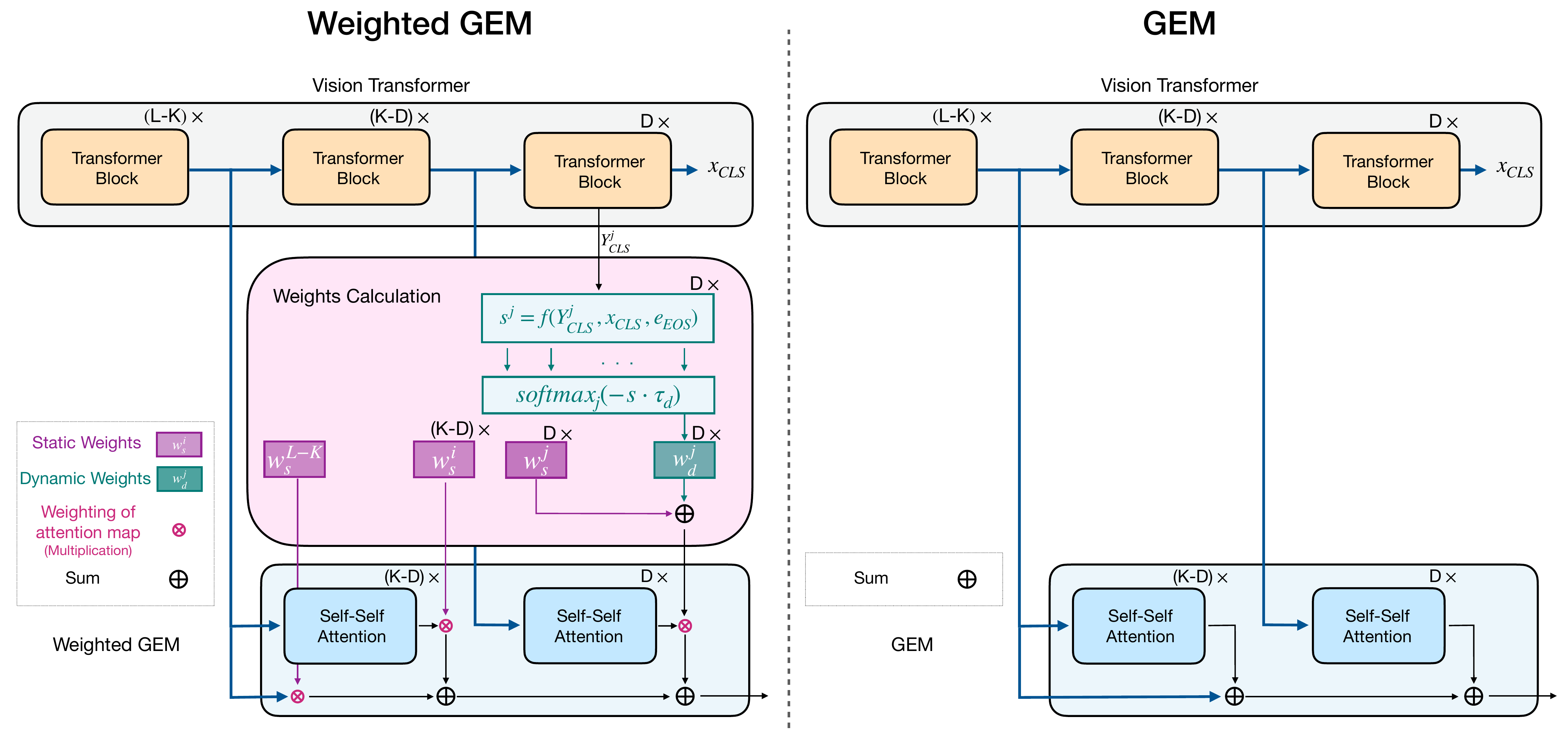}
        \caption{\textbf{Comparison of GEM and our proposed weighting mechanism.} The proposed weighting mechanism is illustrated on the left. Static weights and dynamic weights can be applied independent of each other. While static weights can be set heuristically or via hyperparameter search based on the general pipeline perfromance, dynamic weights are adapting to the importance of the different transformer layers individually and with respect to each prompt as described in \cref{sec:method_LayerWeighting}. Standard GEM does not use any weights, which equals to always using $1-$weights in our formulation.
        }
        \label{fig:gem_comparison}
\end{center}
    \end{figure*}
\section{Weighted GEM vs. GEM}
\label{sec:weighted_gem_vs_gem}
VideoGEM contains two new concepts that we introduced in the main paper: prompt decomposition and a weighting mechanism for GEM. While prompt decomposition has already been illustrated in \cref{fig:teaser,fig:main}, we also want to focus on the difference between GEM and our proposed weighting mechanism for GEM (weighted GEM) in \cref{fig:gem_comparison}. The proposed weighting mechanism is illustrated on the left. Static weights and dynamic weights can be applied independent of each other. While static weights can be set heuristically or via hyperparameter search based on the general pipeline performance, dynamic weights are adapting to the importance of the different transformer layers individually and with respect to each prompt as described in \cref{sec:method_LayerWeighting}. Standard GEM does not use any weights, which equals to always using $1-$weights in our formulation. This results in one weight for the initial self attention output that is the first input into GEM as well as one weight for the output of each self-self attention block. A layer output is multiplied by its corresponding weight, producing the final output as the sum of weighted outputs.
\section{Additional backbone}
\label{sec:blip}
We extend the evaluation of our VideoGEM to another backbone. Namely, we consider the image-text BLIP \cite{li2022blip} model finetuned on the instructional video-text HowToCaption dataset \cite{shvetsova2025howtocaption} (\cref{table:blip_res}). The HowToCaption dataset is based on the HowTo100M \cite{miech2019howto100m} dataset and provides captions for instructional videos. We apply both, the original GEM as well as the proposed VideoGEM to this backbone and evaluate it on on the V-HICO (VH), Daly, YouCook-Interactions (YC), and GroundingYouTube (gYT) datasets. \cref{table:blip_res} shows that also with BLIP as a backbone, VideoGEM outperforms GEM by more than $10\%$ on average, outperforming GEM individually on each dataset. 
Since our proposed prompt decomposition method relies on good predictions for the individual verb, object, and action prompts we assume that the proposed method benefits more from a finetuned backbone with improved individual predictions via prompt decompostion than the original GEM .
Moreover, \cref{table:verb_weights,table:object_weights,table:action_weights} show that BLIP also largely benefits from dynamic and static weights.

\begin{center}
    \begin{table}
        \centering
        \begin{tabular}{ p{1.5cm}|p{0.65cm}p{0.65cm}p{0.65cm}p{0.65cm}|p{0.65cm}  }

         \hline
         Setting & VH & Daly & YC  & gYT & avg\\
         \hline

        GEM & 67.79 & 69.00 &34.77 & 37.97 & 52.38\\

        VideoGEM & \textbf{77.20} & \textbf{72.04} & \textbf{51.5}7 & \textbf{53.83} & \textbf{63.66} \\
        \hline

        \end{tabular}
        \caption{\textbf{Experimental evaluation using the BLIP backbone fine-tuned on the video-text HowToCaption dataset.} We report the model's performance on the V-HICO (VH), Daly, YouCook-Interactions (YC), and GroundingYouTube (gYT) datasets.
        }
        \label{table:blip_res}
    \end{table}
    \vspace{-4mm}
\end{center}

\begin{center}
    \begin{table}
    \resizebox{1\linewidth}{!}{
        \begin{tabular}{ p{1.4cm}p{0.3cm}p{0.5cm}|p{0.65cm}p{0.65cm}p{0.65cm}p{0.65cm}||p{0.65cm}  }

         \hline
         Model & Set & Data & VH & Daly & YC  & gYT & avg\\
         \hline

        \multirow{4}{*}{OpenCLIP}
        & \multirow{2}{*}{base} & img & \textbf{68.28} & \textbf{74.05} & \textbf{56.87} & \textbf{32.91} & \textbf{58.03} \\
         & & vid & 65.20 & 70.68 & 48.67 & 28.15 & 53.18 \\
         \cline{2-8}
         
         & \multirow{2}{*}{ours} & img & \textbf{76.42} & \textbf{80.32} & \textbf{60.05} & \textbf{45.33} & \textbf{65.53} \\
         &  & vid & 74.26 & 76.55 & 52.93 & 42.10 & 61.46\\
         \hline

        \multirow{4}{*}{CLIP}
         & \multirow{2}{*}{base} & img & \textbf{67.79} & \textbf{78.52} &\textbf{50.08} &\textbf{36.92} &\textbf{58.33} \\
         & & vid & 67.55 & 76.31 & 46.10 & 34.82 & 56.20\\
         \cline{2-8}
         
         & \multirow{2}{*}{ours} & img & \textbf{76.90} &\textbf{84.53} &\textbf{52.57} &\textbf{47.46} &\textbf{65.37} \\
         & & vid & 75.15 & 82.68 & 50.36 & 45.79 & 63.50\\
        \hline
        
        \end{tabular}
}
        \caption{\textbf{Experimental evaluation of extending image-text backbones to video input.} CLIP and OpenCLIP are extended to video input in a training-free manner. Their accuracy with image and video data as input is compared for VideoGEM (ours) and standard GEM (base) on the V-HICO (VH), Daly, YouCook-Interactions (YC), and GroundingYouTube (gYT) datasets.}
        \label{table:img_to_vid_abl}
        \vspace{-4mm}
    \end{table}
\end{center}
\section{Video input for image-language models}
\label{sec:video}
Since image-language models, such as CLIP and OpenCLIP, are limited to processing single images as input, we conduct additional experiments where we extend these models to handle video input.

\myparagraph{Setup.}
Let $F=\{f_1, ..., f_T\}$ be a video with $T$ frames. For an input frame $f_i$, the first layer $l_0$ of a backbone encodes the input image into $N$ patch tokens, before they are jointly processed through the transformer layers.
In order to utilize video data we apply the first layer $l_0$ individually to all frames $f_i, 1 \leq i \leq T$ resulting in $T*N$ patch tokens.
Since the transformer layers are independent of the number of patch tokens, the $T*N$ patch tokens are processed by the transformer layers without any architectural change.
For a given input sequence, the patch tokens from all frames are concatenated into a single sequence $X \in \mathcal{R}^{(T*N) \times d}$, where $d$ is the embedding dimension and further processed by GEM as described in \cref{sec:method_videogem}.

\myparagraph{Results.}
We evaluate the performance of GEM and VideoGEM with CLIP and OpenCLIP as backbones for video input with $T=8$ frames compared to image input in \cref{table:img_to_vid_abl}. The frame sampling for video input is the same as for ViCLIP in previous experiments according to \cref{sec:eval_setup}.
Independent of the backbone or the setting (GEM, or VideoGEM) video input decreases the accuracy compared to image input. We attribute this to the fact, that the backbones are not trained to handle video input and therefore are not capable of utilizing the spatial and temporal relationships that are inserted into the input data by using videos, but are distorted by the different input instead. This assumption is supported by the results for ViCLIP in \cref{table:viclip_results}. ViCLIP has a similar architecture to CLIP besides its first embedding layer which embeds a video input into patch tokens. ViCLIP is specifically trained with video input and its accuracy benefits from using videos as input as shown in \cref{table:viclip_results}, suggesting that video-specific pretraining is needed to leverage the spatial and temporal relationships in videos.
\section{Ablation for static and dynamic weights}
\label{sec:weights_abl_suppl}
In this section, we extend our analysis of the effects of static, dynamic, and combined weights (elaborating on the experimental evaluation in \cref{sec:eval_ablation}).
We investigate their effects separately for verb prompts (\cref{table:verb_weights}), object prompts (\cref{table:object_weights}), and action prompts (\cref{table:action_weights}). Note, that compared to \cref{table:weights} in the main paper, \cref{table:verb_weights,table:object_weights,table:action_weights} don't apply prompt decomposition but only the weighting strategies. We include ViCLIP on video and image data, as well as CLIP, OpenCLIP, and BLIP which is finetuned on HowToCaption, as backbones.
First, we observe that independent of the prompt (verb, object, or action) averaged over datasets and backbones, dynamic weights and static weights improve over using no weights. The combination of static and dynamic weights further improves over both, static weights or dynamic weights on their own, showing the capabilities of static and dynamic weights as well as their additive effects. Moreover, OpenCLIP and BLIP improve around $0.5-1\%$ with dynamic weights across different prompts, while the accuracy of ViCLIP and CLIP only shows minor changes. This confirms the assumption that the benefit of dynamic weights highly depends on the backbone. 
\begin{center}
    \begin{table}
    \resizebox{1\linewidth}{!}{
        \begin{tabular}{ p{1.25cm}p{0.4cm}p{0.5cm}|p{0.65cm}p{0.65cm}p{0.65cm}p{0.65cm}||p{0.65cm}  }
        
         \hline
         Model & Data &  Set & VH & Daly & YC & gYT & avg\\
         \hline
         \multirow{4}{*}{ViCLIP} & \multirow{4}{*}{vid} & base & 63.33 & 75.62 &  33.84 & 29.40 & 50.55\\
         && dyn & 63.27 & 75.82&  34.00 & 29.62 & 50.68\\
           & &stat & 64.66 & 76.20& 36.17 & 30.96 & 52.00\\
           & &s+d & \textbf{64.72}  & \textbf{76.21} & \textbf{36.54}&  \textbf{31.26} & \textbf{52.18}\\
         \hline
        
           \multirow{4}{*}{ViCLIP} & \multirow{4}{*}{img} & base & 64.17 & 75.87 & 32.64 & 28.40 & 50.27\\
             & & dyn & 64.17 & 76.03 &32.60 &  28.62 & 50.36\\
           && stat & \textbf{65.44} & \textbf{76.35} & 34.16 & 29.67 & 51.41\\
           && s+d & 65.38 & 76.32 & \textbf{34.24} & \textbf{30.00} & \textbf{51.49}\\
         \hline

           \multirow{4}{*}{CLIP} & \multirow{4}{*}{img} & base & 69.90 & 79.71 & 29.30 & 23.29 & 50.55\\
          && dyn & 70.14 & 79.79 & 29.14 & 22.57 & 50.41 \\
           && stat & 72.32 & 82.06 &  \textbf{30.43} & \textbf{26.44} & \textbf{52.81}\\
           && s+d & \textbf{72.80} & \textbf{82.48} & 29.74 & 25.69 & 52.68\\
         \hline
        
           \multirow{4}{*}{OpenCLIP} & \multirow{4}{*}{img} & base & \textbf{70.87} & 78.65 & 41.16 & 17.75 & 52.11 \\
         && dyn & 70.57 & 78.42 &  41.72 & 18.43 & 52.29\\
           && stat & 69.36 & \textbf{80.69} & 43.09 & 20.59 & 53.43\\
           && s+d & 69.60 & 80.24 &  \textbf{43.45} & \textbf{21.60} & \textbf{53.72} \\
         \hline
           \multirow{4}{*}{BLIP} & \multirow{4}{*}{img} & base & 63.63& 64.75 & 33.00 & 31.18 & 48.17\\
          & &dyn & 64.66 & 65.45 & 33.20 & 31.41 & 48.68\\
           && stat & 68.15 & \textbf{68.23} &  44.86 & 40.41 & 55.41\\
           && s+d & \textbf{69.00} & 67.91 &  \textbf{45.54} &\textbf{40.49} & \textbf{55.74}\\
         \hline
         \hline
        \end{tabular}
        }
        \caption{\textbf{Ablation study on the effect of weights for verb prompts.} Accuracy for verb prompts  for vanilla GEM (base), dynamic weights for the last three layers (dyn),  static weights (stat), and the proposed combination of static and dynamic weights (s+d), evaluated on the V-HICO (VH), Daly, YouCook-Interactions (YC), and GroundingYouTube (gYT) datasets.}
        \label{table:verb_weights}
        \vspace{-4mm}
    \end{table}
\end{center}
\begin{center}
    \begin{table}
        \resizebox{1\linewidth}{!}{
        \begin{tabular}{ p{1.25cm}p{0.4cm}p{0.5cm}|p{0.65cm}p{0.65cm}p{0.65cm}p{0.65cm}||p{0.65cm}  }

         \hline
         Model & Data & Set & VH & Daly & YC & gYT & avg \\
         \hline
        
          \multirow{4}{*}{ViCLIP} & \multirow{4}{*}{vid} & base & \textbf{62.48} & 66.44 &  53.70 & 45.72 & 57.09\\
         & & dyn & 62.00 & 66.53 &  53.82 & 45.75 & 57.03\\
          & & stat & 62.24 & \textbf{68.73} &  54.86 & 46.56 & 58.10\\
           && s+d & 62.30 & 68.71 & \textbf{54.94} & \textbf{46.78} & \textbf{58.18} \\
         \hline
         
           \multirow{4}{*}{ViCLIP} & \multirow{4}{*}{img} & base & 62.06 & 66.83 &  53.86 & 44.39 & 56.81\\
         & & dyn & \textbf{62.18} & 66.92 & 53.98 & 44.63 & 56.93 \\
           && stat & 60.92 & 68.59 &  54.14 & 45.03 & 57.17 \\
           && s+d & 61.10 & \textbf{68.62} & \textbf{54.30} & \textbf{45.17} & \textbf{57.30}\\
         \hline
        
           \multirow{4}{*}{CLIP} & \multirow{4}{*}{img} & base & 65.74 & 76.27 &  46.50 & 36.24 & 56.19\\
         & & dyn & \textbf{66.47} & 76.78 &  46.02 & 35.70 & 56.24\\
           && stat  & 65.68 & 78.70 &  \textbf{47.35} & \textbf{38.06} & \textbf{57.45}\\
           && s+d & 65.56 & \textbf{79.45} & 47.19 & 37.52 & 57.43\\
         \hline
        
           \multirow{4}{*}{OpenCLIP} & \multirow{4}{*}{img} & base & 67.55 & 76.43 & 49.68 & 28.35 & 55.50\\
         & & dyn & \textbf{68.52} & 75.67 & 50.76 & 30.91 & 56.47\\
           && stat & 64.90 & \textbf{77.86} & 50.96 & 32.03 & 56.44\\
           && s+d & 65.44 & 77.00 & \textbf{51.21} &  \textbf{34.20} & \textbf{56.96}\\
         \hline
        
           \multirow{4}{*}{BLIP} & \multirow{4}{*}{img} & base & 59.95 & \textbf{59.85} & 32.68 & 34.17 & 46.66\\
         & & dyn & 60.62 & 59.59 & 33.76 & 35.57 & 47.39\\
           && stat & \textbf{62.48} & 58.09 & 45.34 & 40.27 & 51.55 \\
           && s+d & 62.06 & 57.94 & \textbf{46.38} & \textbf{42.15} & \textbf{52.13}\\
         \hline
         \hline

        
        \end{tabular}
        }
        \caption{\textbf{Ablation study on the effect of weights for object prompts.} Accuracy for object prompts for vanilla GEM (base), dynamic weights for the last three layers (dyn),  static weights (stat), and the proposed combination of static and dynamic weights (s+d), evaluated on the V-HICO (VH), Daly, YouCook-Interactions (YC), and GroundingYouTube (gYT) datasets. }
        \label{table:object_weights}
    \end{table}
    \vspace{-4mm}
\end{center}
\begin{center}
    \begin{table}
    \resizebox{1\linewidth}{!}{
        \begin{tabular}{ p{1.25cm}p{0.4cm}p{0.5cm}|p{0.65cm}p{0.65cm}p{0.65cm}p{0.65cm}||p{0.65cm}  }

         \hline
         Model & Data & Set & VH & Daly & YC  & gYT & avg\\
         \hline
        
          \multirow{4}{*}{ViCLIP} & \multirow{4}{*}{vid} & base & 65.08 & 73.75 &  \textbf{53.62} & 51.28 & 60.93\\
         & & dyn & 64.84 & 73.81  & \textbf{53.62} & 51.25 & 60.88\\
           && stat & \textbf{66.53} & \textbf{74.23}  & 52.97 & 51.93 & \textbf{61.42}\\
           && s+d & 65.68 & 74.17  & 52.97 & \textbf{51.99} & 61.20\\
         \hline

          \multirow{4}{*}{ViCLIP} & \multirow{4}{*}{img} & base & 65.20  & 74.00  & 52.17 & 48.80 & 60.04 \\
         & & dyn & 64.66 & 74.04 & 52.37 & 48.91 & 60.00\\
          & & stat & \textbf{65.86} & \textbf{74.62} & 52.65& 49.51 & 60.66\\
          & & s+d & 65.80 & 74.50  & \textbf{52.85} & \textbf{49.66} & \textbf{60.70}\\
         \hline

           \multirow{4}{*}{CLIP} & \multirow{4}{*}{img} & base & 67.79 & 78.52  & 50.08 & 36.92 & 58.33\\
         & & dyn & 67.43 & 78.47 & 49.68 & 36.83 & 58.10\\
          & & stat & 68.76 & \textbf{80.91} & \textbf{51.37} & \textbf{39.65} & \textbf{60.17} \\
          & & s+d & \textbf{68.94}  & 80.53  & 50.84 & 39.31 & 59.91\\
         \hline
        
           \multirow{4}{*}{OpenCLIP} & \multirow{4}{*}{img} & base & 68.28 & 74.05 & 56.87 & 32.91 & 58.03\\
         & & dyn & \textbf{68.64} & 74.24 & 56.47 & 35.58 & 58.73\\
          & & stat & 66.10 & \textbf{75.50} & \textbf{58.68} & 36.18 & 59.12\\
          & & s+d & 66.77 & 74.87 & 57.36 & \textbf{38.38} & \textbf{59.35}\\
         \hline
        
           \multirow{4}{*}{BLIP} & \multirow{4}{*}{img} & base & 67.79 & 69.00 & 34.77 & 37.97 & 52.38\\
         & & dyn & 69.00 & 68.86 & 37.02 & 39.47 & 53.59 \\
          & & stat & 69.30 & \textbf{71.86} & 46.26 & 45.52 & 58.24\\
          & & s+d & \textbf{70.02} & 70.68 & \textbf{47.99} &  \textbf{46.53} & \textbf{58.81} \\
         \hline
         \hline

        
        \end{tabular}
        }
        \caption{\textbf{Ablation study on the effect of weights for action prompts.} Accuracy for action prompts with vanilla GEM (base), dynamic weights for the last three layers (dyn),  static weights (stat), and the proposed combination of static and dynamic weights (s+d), evaluated on the V-HICO (VH), Daly, YouCook-Interactions (YC), and GroundingYouTube (gYT) datasets.}
        \label{table:action_weights}
    \end{table}
    \vspace{-4mm}
\end{center}
\newcolumntype{P}[1]{>{\centering\arraybackslash}p{#1}}

\begin{table}
\centering
    \begin{tabular}{ 
    P{2.1cm}|P{0.65cm}P{0.65cm}P{0.65cm}P{0.65cm}|P{0.65cm}  }
    
     \hline
     Setting & VH & Daly & YC & gYT & avg \\
     \hline

        W1 & 73.52 & 77.27 & \textbf{56.55} & 56.43 & 65.94 \\
        W2 & 75.57 & 78.38 & 52.37 & 54.99 & 65.33 \\
        W3 & \textbf{77.80} & \textbf{78.74} & 52.33 & 54.98 & 65.96 \\
        org & 75.75 & 78.25 & 55.10 & \textbf{57.21} & \textbf{66.58} \\

       \hline
    \end{tabular} 
\caption{\textbf{Influence of different weighting schemes}. VideoGEM with prompt decomposition and combined weights is evaluated with different weighting schemes: $w_{verb}=0.1, w_{obj}=0.3, w_{act}=0.6$ (W1), $w_{verb}=0.3, w_{obj}=0.1, w_{act}=0.6$ (W2), $w_{verb}=\frac{1}{3}, w_{obj}=\frac{1}{3}, w_{act}=\frac{1}{3}$(W3), $w_{verb}=0.2, w_{obj}=0.2, w_{act}=0.6$ (org). Where \textit{W1} prioritizes objects, \textit{W2} prioritizes verbs, \textit{W3} treats all prompts equally, and \textit{org} are the original weights employed in the main paper. The weighting schemes are evaluated with ViCLIP on video input on V-HICO (VH), Daly, YouCook-Interactions (YC), and GroundingYouTube (gYT).}
\label{table:weighting_abl}
\vspace{-4mm}
\end{table}
\section{Ablation for weights}
For prompt decomposition the same weights of $w_{verb}=0.2, w_{obj}=0.2, w_{act}=0.6$ are applied in all experiments. To analyze the importance of the weighting scheme we compare $4$ different weighting schemes in \cref{table:weighting_abl}. One that prioritizes objects (W1), one that prioritizes verbs (W2), one that treats all three prompts equally (W3), and our original weighting scheme (org). We observe similar accuracy for the different weighting schemes, while our original weights perform the best.
\newcolumntype{P}[1]{>{\centering\arraybackslash}p{#1}}

\begin{table}
\centering
    \begin{tabular}{ 
    P{2.1cm}|P{0.65cm}P{0.65cm}P{0.65cm}P{0.65cm}|P{0.65cm}  }
    
     \hline
     Model & VH & Daly & YC & gYT & avg \\
     \hline

        Qwen-VL & \textbf{84.20} & 63.05 & 28.49 & \textbf{57.69} & 58.36 \\
        VideoGEM & 75.74 & \textbf{78.25} & \textbf{55.10} & 57.21 & \textbf{66.58} \\

       \hline
    \end{tabular} 
\caption{\textbf{Comparison to multimodal LLMs}. VideoGEM is compared to a multimodal large language model, namely Qwen-VL \cite{bai2023qwenvlversatilevisionlanguagemodel}. For Qwen-VL the prompt template "Generate grounding for {}" is applied. The center of the predicted bounding box is the prediction of Qwen-VL. VideoGEM is applied with ViCLIP as a backbone on video input. Both models are evaluated on V-HICO (VH), Daly, YouCook-Interactions (YC), and GroundingYouTube (gYT).}
\label{table:qwen_abl}
\vspace{-4mm}
\end{table}
\section{Comparison to multimodal LLMs}
We extend our comparison to State-of-the-art methods to multimodal large language models, namely Qwen-VL \cite{bai2023qwenvlversatilevisionlanguagemodel}. Note that Qwen is not directly comparable, as it utilizes location information during training and differs significantly in both the number of parameters and the amount of training data. We provide a comparison of Qwen-VL and VideoGEM in \cref{table:prompt_merg}. We use "Generate grounding for {}" as a prompt template for Qwen-VL. We use a different prompt for Qwen-VL compared to VideoGEM to ensure the highest rate of bounding box predictions for Qwen-VL. The center of the predicted bounding box is the prediction for Qwen-VL. The results indicate that our training-free method VideoGEM outperforms Qwen-VL on average.
\newcolumntype{P}[1]{>{\centering\arraybackslash}p{#1}}

\begin{table}
\centering
    \begin{tabular}{ 
    P{2.1cm}|P{0.65cm}P{0.65cm}P{0.65cm}P{0.65cm}|P{0.65cm}  }
    
     \hline
     Setting & VH & Daly & YC & gYT & avg \\
     \hline

        avg. txt. & 63.33 & 63.10 & 51.09 & 54.37 & 57.97 \\
        avg. heat. & 63.33 & 63.04 & 51.09 & 54.41 & 57.97 \\
        VideoGEM & \textbf{75.75} & \textbf{78.25} & \textbf{55.10} & \textbf{57.21} & \textbf{66.58} \\

       \hline
    \end{tabular} 
\caption{\textbf{Comparison to CLIP prompting}. The prompt decomposition of VideoGEM is compared to the prompting technique of CLIP. The $80$ prompt templates that are provided by the CLIP paper are used to generate $80$ prompts. These prompts are used in two different settings. In the first setting, the $80$ prompts are averaged, resulting in one text embedding. The combined weights of VideoGEM are then applied (\textit{avg. txt.}). In the second setting, the combined weights are applied to each prompt individually resulting in $80$ heatmaps (\textit{avg. heat.}). The final prediction in both settings is obtained by taking the location of the maximum logit in the heatmap.
VideoGEM is applied with ViCLIP as a backbone on video input. Both models are evaluated on V-HICO (VH), Daly, YouCook-Interactions (YC), and GroundingYouTube (gYT).}
\label{table:clip_abl}
\vspace{-4mm}
\end{table}
\section{Ablation for CLIP prompting}
We compare the prompt decomposition of VideoGEM with the prompting technique proposed by CLIP in \cref{table:clip_abl}. We apply the $80$ prompt templates that are provided by the CLIP paper to the test labels. Two different settings are used to obtain the final prediction. In the first setting, the text embeddings are averaged (\textit{avg. txt.}) and then the combined weights of VideoGEM are applied to the single averaged text embedding resulting in a single heatmap. In the second setting, the combined weights are applied to each of the $80$ prompts individually resulting in $80$ heatmaps that are averaged pointwise (\textit{avg. heat.}).  The location of the maximum logit in the heatmap is the predicted location for both settings, similar to VideoGEM.
VideoGEM significantly outperforms CLIPs prompting technique. This shows, that the benefit of our proposed prompt decomposition comes not only from the majority voting but the prompt decomposition itself.
\section{Qualitative analysis}
\label{sec:qual_suppl}
\begin{center}
    \begin{figure*}
        \includegraphics[width=0.99\linewidth]{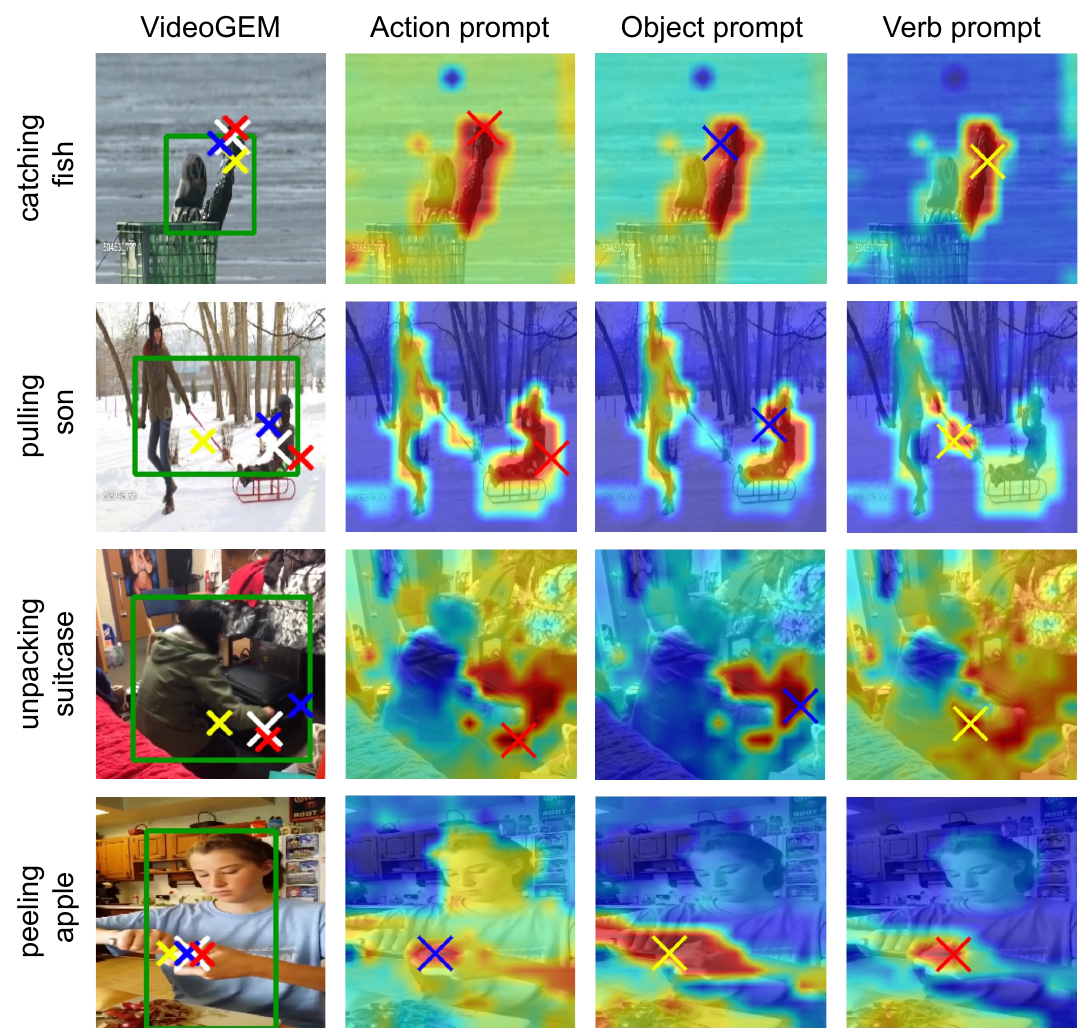}
        \caption{\textbf{Qualitative examples for VideoGEM on V-HICO}. VideoGEM is applied with ViCLIP on video data. The main frame with its label is shown. The green bounding box is the ground truth and the white cross is the final prediction of VideoGEM. Besides that are the heatmaps for the action, object, and verb prompt. The individual predictions for the action, object, and verb prompt are shown by red, blue, and yellow crosses respectfully. The ground truth label of the image is shown on the left.
        }
        \label{fig:vhico_ensemble}
    \end{figure*}
\end{center}

\begin{center}
    \begin{figure*}
        \includegraphics[width=0.99\linewidth]{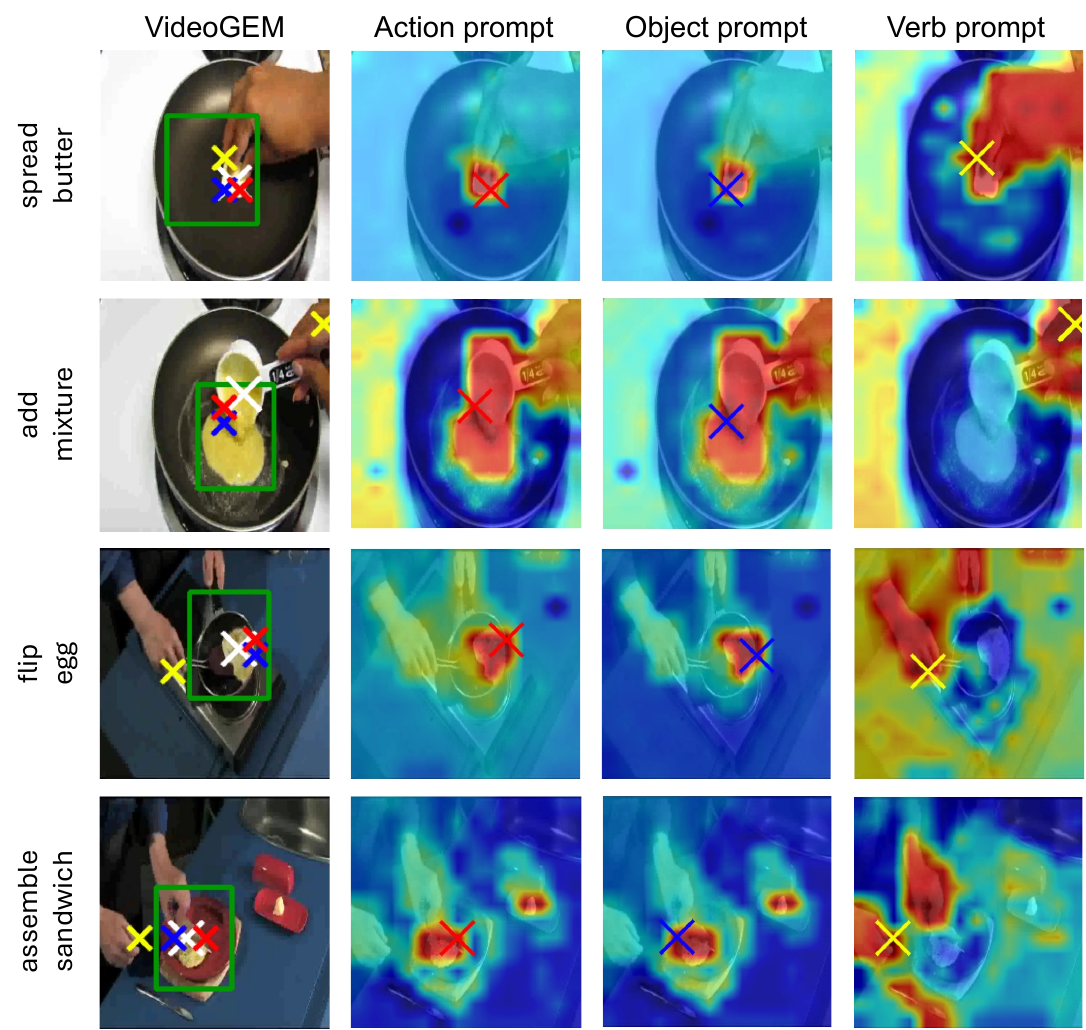}
        \caption{\textbf{Qualitative examples for VideoGEM on GroundingYouTube}. VideoGEM is applied with ViCLIP on video data. The main frame with its label is shown. The green bounding box is the ground truth and the white cross is the final prediction of VideoGEM. Besides that are the heatmaps for the action, object, and verb prompt. The individual predictions for the action, object, and verb prompt are shown by red, blue, and yellow crosses respectfully. The ground truth label of the image is shown on the left.}
        \label{fig:grYT_ensemble}
    \end{figure*}
\end{center}

We present qualitative examples of predictions from our VideoGEM model, using ViCLIP as the backbone with video inputs on the  V-HICO and GroundingYouTube datasets in \cref{fig:grYT_ensemble,fig:vhico_ensemble} respectively.
The ground truth bounding box is green and the final VideoGEM prediction is white. The individual prompt predictions and heatmaps are shown for action, object, and verb prompts in red, blue, and yellow respectively. We observe that the heatmaps and the predicted locations for verb prompts differ the most from action and object prompts. Mostly all three predicted locations are close together, especially for actions with small spatial scale such as \textit{``spread butter''}. However, when the action is bigger as for \textit{``unpacking suitcases''} VideoGEM centers the action between its components. Moreover, if predictions are slightly off, such as in \textit{``catching fish''}, or \textit{``pulling son''}, where the action prompt initially is outside the ground truth bounding box focusing only on the object, 
VideoGEM drags the prediction back into the bounding box, leading to more robust and accurate predictions.

\section{Limitations}
\label{sec:limitations}
VideoGEM is designed for spatial action grounding. It considers the temporal context of videos only implicitly via its video backbone. VideoGEM's predictions are spatial locations for the given input images/videos without temporal predictions. Moreover, the proposed prompt decomposition only works for action grounding in its current state. However, this could be adapted to more general settings.


\end{document}